\documentclass[wcp]{jmlr}

\usepackage{longtable}

\usepackage{booktabs}

\title[Radical-level Ideograph Encoder]{Radical-level Ideograph Encoder for RNN-based Sentiment Analysis of Chinese and Japanese}

  \author{\Name{Yuanzhi Ke} \Email{enshika8811.a6@keio.jp}\\
  \addr Department of Information and Computer Science, 
  Faculty of Science and Engineering, Keio University
  \AND
  \Name{Masafumi Hagiwara} \Email{hagiwara@keio.jp}\\
  \addr Department of Information and Computer Science, 
  Faculty of Science and Engineering, Keio University
 }

\editors{Yung-Kyun Noh and Min-Ling Zhang}

\begin{document}

\maketitle

\begin{abstract}
The character vocabulary can be very large in non-alphabetic languages such as Chinese and Japanese, which makes neural network models huge to process such languages. We explored a model for sentiment classification that takes the embeddings of the radicals of the Chinese characters, i.e, hanzi of Chinese and kanji of Japanese. Our model is composed of a CNN word feature encoder and a bi-directional RNN document feature encoder. The results achieved are on par with the character embedding-based models, and close to the state-of-the-art word embedding-based models, with 90\% smaller vocabulary, and at least 13\% and 80\% fewer parameters than the character embedding-based models and word embedding-based models respectively. The results suggest that the radical embedding-based approach is cost-effective for machine learning on Chinese and Japanese.
\end{abstract}

\section{Introduction}
Word embeddings have been widely used for natural language processing (NLP) tasks \citep{collobert2011natural}. However, the large word vocabulary makes word embeddings expensive to train. Some people argue that we can model languages at the character-level~\citep{kim2016character}.
For alphabetic languages such as English, where the characters are much fewer than the words, the character embeddings achieved the state-of-the-art results with much fewer parameters.

Unfortunately, for the other languages that use non-alphabetic systems, the character vocabulary can be also large. Moreover, Chinese and Japanese, two of the most widely used non-alphabetic languages, especially contain large numbers of ideographs: hanzi of Chinese and kanji of Japanese. The character vocabulary can be as scalable as the word vocabulary (e.g., see the datasets introduced in Section~\ref{ssec:datasets}). Hence the conventional character embedding-based method is not able to give us a slim vocabulary on Chinese and Japanese.

For convenience, let us collectively call hanzi and kanji as Chinese characters.
Chinese characters are ideographs composed with semantic and phonetic radicals, both of which are available for character recognition, and the semantic information may be embedded in Chinese characters by the semantic radicals~\citep{williams2010chinese}. 
Besides, though the character vocabulary is huge, the number of the radicals is much fewer. 
Accordingly, we explored a model that represents the Chinese characters by the sequence of the radicals. We applied our proposed model to sentiment classification tasks on Chinese and Japanese and achieved the follows:
\begin{itemize}
	\item The results achieved by our proposed model are close to the state-of-the-art word embedding-based models, with approximately 90\% smaller vocabulary, 91\% and 82\% fewer parameters for Chinese and Japanese respectively.
	\item The results are on par with the character embedding-based models, with approximately 13\% fewer parameters, and 90\% smaller vocabulary for both Chinese and Japanese.
\end{itemize}
 

\section{Methodology}

\begin{figure}[htp]
	\begin{center}
		\includegraphics[width=\textwidth]{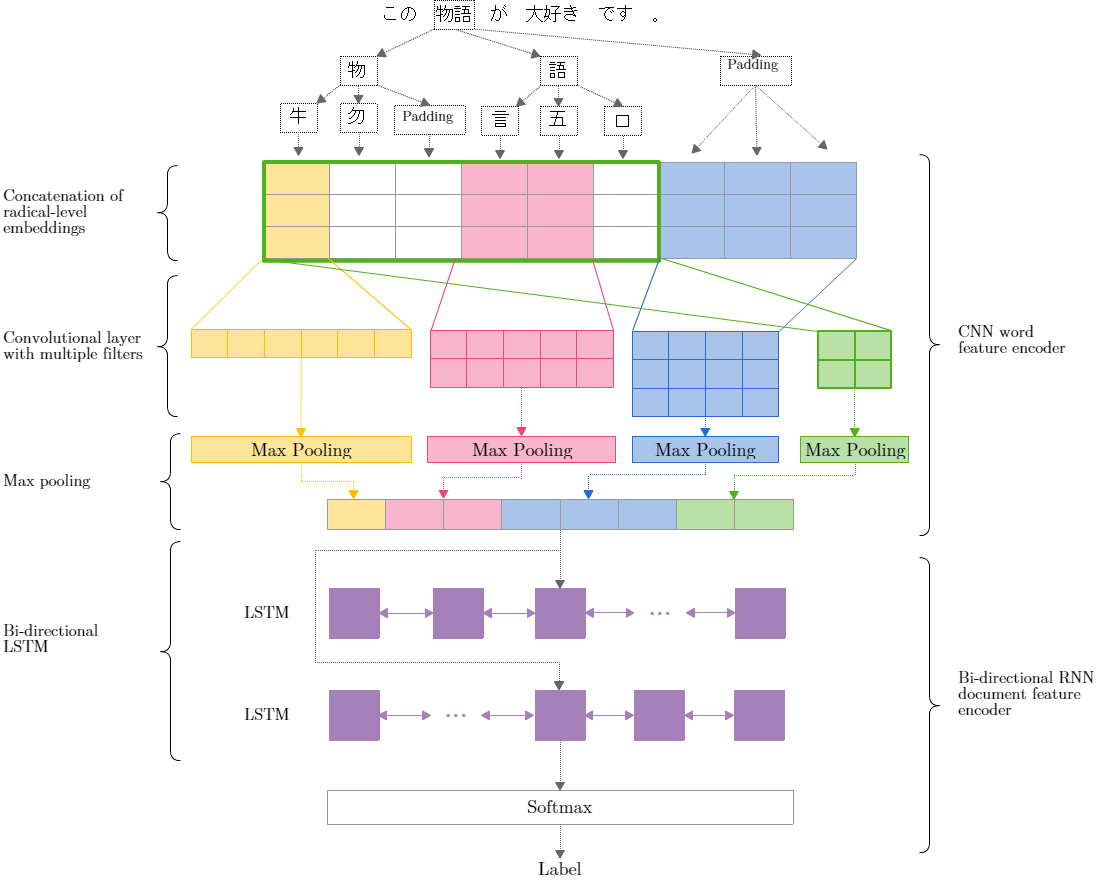}
		\caption{Architecture of our proposed model. The Chinese characters are split into radicals from the left to the right, the top to the bottom. Each character is represented by a sequence of radical-level embeddings. The radical embedding sequences of the characters in a word are input into a convolutional layer with multiple filters to encode the word features, which are then pooled max-over-time and input into a bi-directional RNN. The bi-directional RNN encodes the document-level feature, and after which the estimation of the label is obtained by applying affine transformation and softmax on the document-level feature. In the figure, there are four filters denoted with different colors. Their widths, strides, and numbers of output channels are different. The filters with stride of one are for radical-level information. Those with stride of three are for character-level information.}\label{fig:architecture}
	\end{center}
\end{figure}

The architecture of our proposed model is as shown in Fig.~\ref{fig:architecture}.
It looks similar to the character-aware neural language model proposed by \citet{kim2016character}, but we represent a word by the sequence of radical embeddings instead of character embeddings. Besides, unlike the former model, there are no highway layers in the proposed model, because we find that highway layers do not bring significant improvements to our proposed model (see Section~\ref{ssec:highway_layers}).

\subsection{Representation of Characters: Sequences of Radical-level Embeddings}

For every character, we use a sequence of $n$ radical-level embeddings to represent it. They are not treated as in a bag because the position of each radical is related to how it is informative~\citep{hsiao2007examination}. For a Chinese character, it is the sequence of the radical embeddings. When it comes to the other characters, including kanas of Japanese, alphabets, digits, punctuation marks, and special characters, it is the sequence comprised of the corresponding character embedding and $n-1$ zero vectors. We zero-pad the radical sequences of all the characters to align the lengths.

\subsection{From Character Radicals to Word Features: CNN Encoder}

CNNs \citep{lecun1989backpropagation} have been used for various NLP tasks and shown effective~\citep{collobert2011natural,kim2014,kim2016character}. For NLP, CNNs are able to extract the temporal features, reduce the parameters, alleviate over-fitting and improve generalization ability.
We also take advantage of the weight sharing technology of CNNs to learn the shared features of the characters.

Let $\mathcal{C}$ be the radical-level vocabulary that contains Chinese character radicals, kanas of Japanese, alphabets, digits, punctuation marks, and special characters, $\mathbf{Q} \in \mathbb{R}^{d \times |\mathcal{C}|}$ be the matrix of all the radical-level embeddings, $d$ be the dimension of each radical-level embedding. We have introduced that each character is represented by a sequence of radical-level embeddings of length $n$. Thus a word $k$ composed of $m$ characters is represented by a matrix $\mathbf{C}^k \in \mathbb{R}^{d \times (m \times n)}$, each column of which is a radical-level embedding.

We apply convolution between $\mathbf{C}^k$ and several filters (convolution kernels). For each filter $h$, we apply a nonlinear activation function $g$ and a max-pooling on the output to obtain a feature vector. Let $r$ be the stride, $w$ be the window, $\mathbf{H}_h$ be the hidden weight of a filter, respectively. The feature vector of $k$ obtained by $h$ is given by:
\begin{equation}
\mathbf{x}_h^k = \mathrm{maxpool}(g(\mathbf{C}^k) \star \mathbf{H}_h + \mathbf{b})
\end{equation}
where $\star$ is the convolution operator. The pooling window is $\frac{m \times n}{r}-w+1$ to obtain the most important information of word $k$.

We have two kinds of filters: (1) the filters with stride $r=1$ to obtain radical-level features; (2) the filters with stride $r=n$ to obtain character-level features.

After the max-pooling layer, we concatenate and flatten all of the outputs through all of the filters as the feature vector of the word. 
Let $a$ be the number of the output channel of each filter. If we use totally $h$ filters, each output of which is $\mathbf{x}_h^k = [x_{h1}\ x_{h2}\ x_{h3}\ ...\ x_{ha}]$, the output feature of $k$ is $\mathbf{x}_h^k = [x_{11}\ x_{22}\ x_{33}\ ...\ x_{1a}\ x_{21}\ x_{22}\ x_{23}\ ...\ x_{2a}\ x_{h1}\ x_{h2}\ x_{h3}\ ...\ x_{ha}]$. Here, we assume that the number of the output channels of every filter is the same, but we tailor it for each filter in the experiments following \citet{kim2016character}.

\subsection{From Word Features to Document Features: Bi-directional Long Short-term Memory RNN Encoder}

An RNN is a kind of neural networks designed to learn sequential data. The output of an RNN unit at time $t$ depends on the output at time $t-1$. Bi-directional RNNs~\citep{schuster1997bidirectional} are able to extract the past and future information for each node in a sequence, have shown effective for Machine Translation~\citep{bahdanau2015neural} and Machine Comprehension~\citep{DBLP:journals/corr/KadlecSBK16}.

A Long Short-term Memory (LSTM)~\citep{Hochreiter:1997:LSM:1246443.1246450} Unit is a kind of unit for RNN that keeps information from long range context. We use a bi-directional RNN of LSTM to encode the document feature from the sequence of the word features.

An LSTM unit contains a forget gate $\mathbf{f}_t$ to decide whether to keep the memory, an input gate $\mathbf{i}_t$ to decide whether to update the memory and an output gate $\mathbf{o}_t$ to control the output.
Let $\mathbf{h}_{t}$ be the output of a LSTM unit at time $t$, $\tilde{\mathbf{\gamma}}_{t}$ be the candidate cell state at time $t$, $\mathbf{\gamma}_{t}$ be the cell state at time $t$. They are given by:
\begin{equation}
\begin{split}
&\mathbf{f}_t=\sigma\left (\mathbf{W}_f\left [\mathbf{h}_{t-1}, \mathbf{x}_t\right ] + \mathbf{b}_f\right ) \\
&\mathbf{i}_t=\sigma\left (\mathbf{W}_i\left [\mathbf{h}_{t-1}, \mathbf{x}_t\right ] + \mathbf{b}_i\right ) \\
&\tilde{\mathbf{\gamma}}_t=\tanh\left (\mathbf{W}_c\left [\mathbf{h}_{t-1}, \mathbf{x}_t\right ] + \mathbf{b}_c\right ) \\
&\mathbf{\gamma}_t=\mathbf{f}_t \ast \mathbf{\gamma}_{t-1} + \mathbf{i}_t \ast \tilde{\mathbf{\gamma}}_t \\
&\mathbf{o}_t = \sigma \left ( \mathbf{W}_o \left [ \mathbf{h}_{t-1}, \mathbf{x}_t\right ] + \mathbf{b}_o\right ) \\
&\mathbf{h}_t = \mathbf{o}_t \ast \tanh \left (\mathbf{\gamma}_t \right )
\end{split}
\end{equation}
where, $\sigma\left (\cdot \right )$ and $\ast$ are the element-wise sigmoid function and multiplication operator.

Our proposed model contains two RNN layers that read document data from different directions. Let $s$ be a document composed of $l$ words. One of the RNN layers reads the document from the first word to the $l$th word, the other reads the document from the $l$th word to the first word. Let $\mathbf{h}_l$ be the final output of the former RNN layer and $\mathbf{h}_1$ be the final output of the latter. We concatenate $\mathbf{h}_l$ and $\mathbf{h}_1$ as the document feature. After that, we apply an affine transformation and a softmax to obtain the prediction of the sentiment labels:
\begin{equation}
\text{\textbf{Pr}}\left(\hat{y}=i|s\right) = \frac{exp\left(\mathbf{W}_p^i\mathbf{z}+\mathbf{b}_p^i\right)}{\sum_{i' \in \mathcal{E}} 
	exp \left(\mathbf{W}_p^{i'}\mathbf{z} + \mathbf{b}_p^{i'}\right)}
\end{equation}
where $z=h_l^\frown h_1$, $^\frown$ is the concatenation operator. $\hat{y}$ is the estimated label of the document, $i$ is one of the labels in the label set $\mathcal{E}$. 

We minimize the cross entropy loss to train the model. Let $\mathcal{S}$ be the set of all the documents, and $y$ be the true label of document $s$, the loss is given by:
\begin{equation}
\mathcal{L} = - \sum_{s \in \mathcal{S}}
\left(\log \text{\textbf{Pr}}(\hat{y}=y|s) \right)
\end{equation}

\section{Experimental Setup}
\subsection{Datasets}
\label{ssec:datasets}

In the experiments, we used a Chinese dataset and a Japanese dataset. 
We used the publicly available Ctrip review data pack~\footnote[1]{http://www.datatang.com/data/11936} for Chinese. They are comprised of travel reviews crawled from ctrip.com~\footnote[2]{http://www.ctrip.com}. We used a subset of 10,000 reviews in the pack. We randomly select 8,000 and 2,000 from it for training and test, respectively.
The Japanese dataset is provided by Rakuten, Inc. It contains 64,000,000 reviews of the products in Rakuten Ichiba~\footnote{http://www.rakuten.co.jp/}. The reviews are labeled with 6-point evaluation of 0-5. We labeled the reviews with less than 3 points as the negative samples, and the others as the positive samples. We randomly chose 10,000 reviews to align the size of the Chinese datasets, 8,000 and 2,000 from it for training and test, respectively.

\begin{table}[htb]
	\centering
	\caption{The statistics of the datasets. $\left|\mathcal{S}\right|=$ the number of the samples in each dataset; $\left|\mathcal{C}\right| =$ the size of the radical-level vocabulary; $\left|\mathcal{V}\right| =$ the size of the character vocabulary; $\left|\mathcal{W}\right| =$ the size of the word vocabulary; $T_c =$ the number of Chinese characters; $T_v=$ the number of all kinds of characters; $T_w=$ the number of the words. The radical-level vocabularies contain Chinese characters, other CJK characters, alphabets, digits, punctuation marks and special characters.}
	\label{tab:dataset_stat}
	\begin{tabular}{lcccccccc}
		\hline
		Dataset & $\left|\mathcal{S}\right|$ & $\left|\mathcal{C}\right|$ & $\left|\mathcal{V}\right|$ & $\left|\mathcal{W}\right|$ & $T_c$ & $T_v$ & $T_w$ & Language \\ \hline 
		Ctrip & 10k & 2k & 21k & 30k & 930k & 1,256k & 795k & Chinese \\ 
		Rakuten & 10k & 2k & 21k & 18k & 941k & 1,005k & 545k & Japanese \\ \hline
	\end{tabular}
\end{table}

The detailed information of the datasets is shown in Table~\ref{tab:dataset_stat}. The character vocabularies are as scalable as the word vocabularies but the radical-level vocabularies are much smaller.
The character vocabulary of the Rakuten dataset is even larger than its own word vocabulary. 94\% of the Rakuten dataset are Chinese characters while the Ctrip dataset contains 74\% Chinese characters. 
Chinese characters account for fewer percentage in Ctrip data, probably because the Ctrip data is not well stripped.

\subsection{Baselines}

We compared the proposed model with the follows:
\begin{itemize}
	\item \textbf{The character-aware neural language model}~\citep{kim2016character}: It is an RNN language model that takes character embeddings as the inputs, encodes them with CNNs and then input them to RNNs for prediction. It achieved the state-of-the-art as a language model on alphabetic languages. We let it predict the sentiment labels instead of words.
	\item \textbf{Bi-directional RNN~\citep{schuster1997bidirectional} with word embeddings}: It is a classical bi-directional RNN classifier, basic but effective. We also employed LSTM for it, and input the word embeddings.
	\item \textbf{Hierarchical attention networks}~\citep{yang2016hierarchical}: It is the state-of-the-art RNN-based document classifier. Following their method, the documents were segmented into shorter sentences of 100 words, and hierarchically encoded with bi-directional RNNs.
	\item \textbf{FastText}~\citep{joulin2016bag}: It is the state-of-the-art baseline for text classification, which simply takes n-gram features and classifies sentences by hierarchical softmax. We used the word embedding version but did not use the bigram version because the other models for comparison do not use bigram inputs.
\end{itemize}

\subsection{Hyperparameters}

The setup of the hyperparameters in our experiments is shown in Table~\ref{tab:model_hyperp}. They were tuned on the development set of 4,000 reviews, 2,000 from another subset in the public Ctrip data pack, and the other 2,000 randomly chosen from the review data of Rakuten Ichiba. We aligned the sizes of the feature vectors of the words and the documents in different models for a fair comparison. All the embeddings are initialized randomly with the uniform distribution.

\begin{table}[htb]
	\centering
	\caption{The setup of the hyperparameters tuned for the experiments. $w=$ the filter width; $r=$ the filter stride; $a=$ the number of the output channels, as a function of $\frac{w}{r}$; $g=$ the nonlinear activation function; $d_\mathbf{c}=$ the dimensions of the radical-level embeddings or the character embeddings, in the proposed model and \citet{kim2016character}'s model respectively; $d_\mathbf{x}=$ the dimension of the word embedding, or the output of the CNN encoder in the proposed model and \citet{kim2016character}; $d_\mathbf{z}=$ the dimension of the document feature vector. They were tuned on the development set of totally 4,000 reviews.}
	\label{tab:model_hyperp}
	\begin{tabular}{l|ccccccc}
		\hline
		\multicolumn{8}{l}{Non-word embedding-based models}\\
		\hline
		Model & 
		$w$ & $r$ & $a$ & $d_\mathbf{c}$ & $d_\mathbf{x}$ & $d_\mathbf{z}$& $g$  \\ \hline
		The proposed&$[1,2,3,3,6,9]$ & $[1,1,1,3,3,3]$ & $[50\cdot \frac{w}{r}]$ & 15 & 600 & 300 & ReLU \\
		\hline
		\citet{kim2016character} & $[1,2,3]$ & $[1,1,1]$ & $[100\cdot \frac{w}{r}]$ & 15 & 600 & 300 & ReLU \\
		\hline
		\multicolumn{8}{l}{}\\
		\hline
		\multicolumn{8}{l}{Word embedding-based models}\\
		\hline
		\multicolumn{5}{l|}{Model} & $d_\mathbf{x}$ & $d_\mathbf{z}$ & $g$ \\ \hline
		\multicolumn{5}{l|}{\citet{schuster1997bidirectional}}& 600 & 300&ReLU  \\
		\hline
		\multicolumn{5}{l|}{\citet{yang2016hierarchical}} & 600 & 300&tanh  \\
		\hline
		\multicolumn{5}{l|}{\citet{joulin2016bag}} & 600 & 300&Linear  \\
		\hline
	\end{tabular}
\end{table}

All of the models were trained by RMSprop~\citep{tieleman2012lecture} with mini-batches of 100 samples. The learning rate and decay term were set as 0.001 and 0.9 respectively, also tuned on the development set. 

\subsection{Text Preprocess}
\label{ssec:text_preprocess}

We segmented the documents into words by Jieba~\footnote{https://github.com/fxsjy/jieba} and Juman++\citep{morita2015morphological}~\footnote{For some non-Japanese tokens that creep in the dataset, Juman++ throws errors In such cases, we used Janome (http://mocobeta.github.io/janome/) instead.}, respectively for Chinese and Japanese. We zero-padded the length of the sentences, words, and radical sequences of the characters as 500, 4 and 3, respectively.

\begin{figure}[htp]
	\begin{center}
		\includegraphics[width=0.4\textwidth]{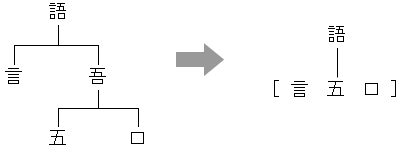}
		\caption{We split the Chinese characters into radicals from the left to the right, the top to the bottom.}\label{fig:char_split}
	\end{center}
\end{figure}

We split the Chinese characters in CJK Unified Ideographs of ISO/IEC 10646-1:2000 character set, until there is no component can be split further, according to CHISE Character Structure Information Database~\footnote{http://www.chise.org/ids/}. Then the Chinese character is represented by the sequence of the radicals from the left to the right, the top to the bottom as shown in Fig.~\ref{fig:char_split}. The sequences are zero-padded to the same length. For an unknown Chinese character not in the set, we treat it as a special character.

\section{Results}

\begin{figure}[htp]
	\begin{center}
		\includegraphics[width=\textwidth]{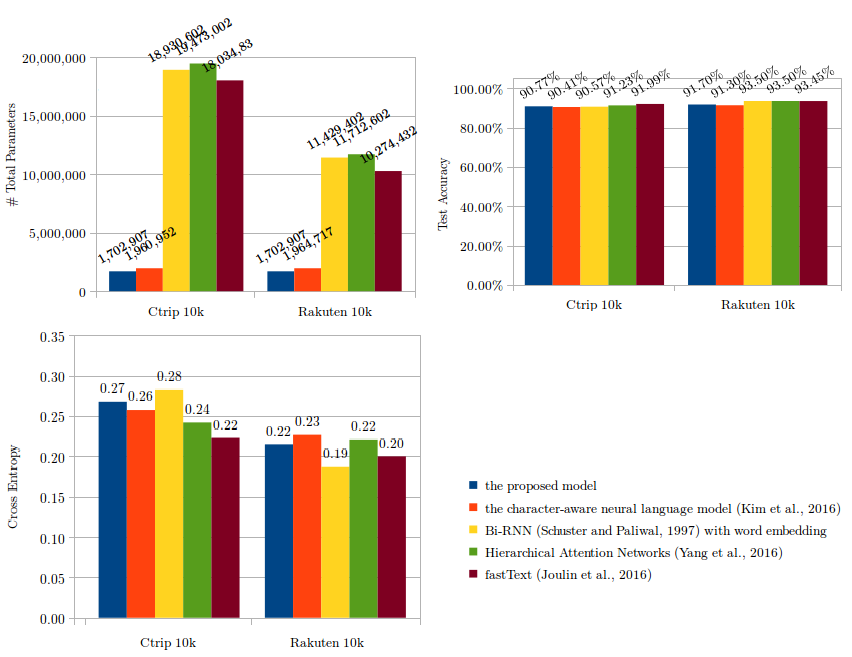}
		\caption{The number of parameters, test accuracy and cross entropy loss off each model. The proposed model has approximately 13\% fewer parameters than the character embedding-based model, 91\% and 82\% fewer parameters than the word embedding-based models for Ctrip dataset and Rakuten dataset respectively. The performance of the proposed model is statistically the same as the character embedding-based model for all the datasets, approximately 99\% of the word embedding-based model for Ctrip dataset, and 98\% of the word embedding-based model for Rakuten dataset, with fewer parameters. The cross entropy losses are also close, especially on Rakuten dataset. The models are close to each other while the proposed model has the fewest parameters.}\label{fig:r3}
	\end{center}
\end{figure}

The number of parameters, test accuracy, and cross entropy loss of each model are as shown in Fig.~\ref{fig:r3}. The proposed model has 13\% fewer parameters than the character embedding-based model, 91\% and 82\% fewer parameters than the word embedding-based models for Ctrip dataset and Rakuten dataset, respectively. The accuracy is statistically the same as the character embedding-based model, approximately 98\% of the word embedding-based model. The losses of the models are also close. The hierarchical attention networks and fastText achieved approximately 11\% and 19\% lower loss on Ctrip dataset. But on Rakuten dataset whose percentage of Chinese characters is higher, the differences between them and the proposed model drops to 0\% and 9\% respectively.

\section{Discussions}

\subsection{The Proposed Model Is the Most Cost-effective} 
The performance of the proposed model is not significantly different from the character embedding-based baseline, and very close to the word embedding-based baselines, with a smaller vocabulary and fewer parameters. It indicates that radical-embeddings are at least as effective as the character-embeddings for Chinese and Japanese, but require less space. It suggests that for Chinese and Japanese, the radical embeddings are more cost-effective than the character embeddings.

\subsection{The CNN Encoder Is Efficient} Even though the character vocabulary is as scalable as the word vocabulary on Chinese and Japanese, the character embedding-based method with CNN encoder can still reduce approximately 90\% and 80\% parameters for the Chinese and Japanese datasets, respectively. The CNNs allow low-dimension inputs, and share weights in the procedure of encoding the intpus to the high-dimension word features. It is probably the reason that it can save parameters although the sizes of the vocabularies are similar.


\subsection{Highway Layers Are Not Effective For Us}
\label{ssec:highway_layers}

\begin{figure}[htp]
	\begin{center}
		\includegraphics[width=\textwidth]{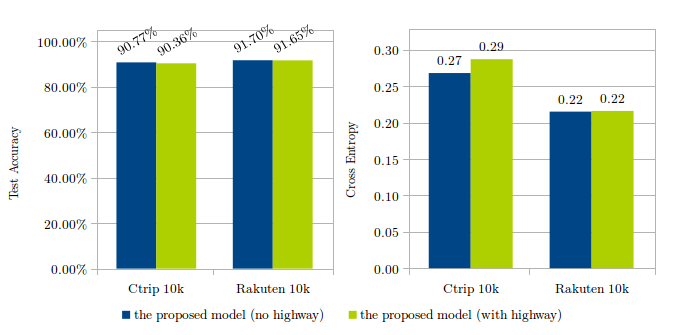}
		\caption{The test accuracy and cross-entropy loss with and without highway layers. No indicated difference is observed.}\label{fig:highway}
	\end{center}
\end{figure}

\citet{kim2016character} reported that the highway networks~\citep{Srivastava:2015:TVD:2969442.2969505} are effective for RNN language models. A highway layer is tailored to adaptively switch between a full-connected layer and a ``highway" that directly outputs the input. We also studied that whether it is effective for our proposed model in the sentiment classification task. Following \citet{kim2016character}, we attempted to input the flattened concatenated output of the max-pooling layer to a highway layer that employs ReLU before we input it to RNN.
The change of the performance is as shown in Fig.~\ref{fig:highway}.

We observed no significant improvement. Probably for two-class sentiment classification, a full-connected layer with ReLU is not necessary between the CNN encoder and the bi-directional RNN encoder, hence the highway network learned to pass the inputs directly to the outputs all the time.

\section{Related Works}

The computational cost brought by the large word vocabulary is a classical problem when neural networks are employed for NLP.
In the earliest works, people limited the size of the vocabulary, which is not able to exploit the potential generalization ability on the rare words~\citep[Chapter~12]{Goodfellow-et-al-2016}. It has made people explore alternative methods for the softmax function to efficiently train all the words, e.g., hierarchical softmax~\citep{morin2005hierarchical}, noise-contrastive estimation~\citep{mnih2012fast} and negative sampling~\citep{mikolov2013distributed}. 
However, the temporal complexity of the softmax function is not the only thing suffering the high-dimension vocabulary. Scalable word vocabulary leads to a large embedding layer, hence huge neural network with millions of parameters, which costs quite a few gigabytes to store.
\citet{zhang2015character} proposed a convolutional neural network (CNN) that takes characters as the input for text classification and outperforms the previous models for large datasets. They showed the character-level CNNs are effective for text classification without the need for words. \citet{kim2016character} introduced a recurrent neural network (RNN) language model that takes character embeddings encoded by convolutional layers as the input. Their model has much fewer parameters than the models using word embeddings, and reached the performance of the state-of-the-art on English, and outperformed baselines on morphologically rich languages. However, for Chinese and Japanese, the character vocabulary is also large, and the character embeddings are blind to the semantic information of the radicals.

\section{Conclusion and Outlook}

We have proposed a model that takes radicals of characters as the inputs for sentiment classification on Chinese and Japanese, whose character vocabulary can be as scalable as word vocabulary. Our proposed model is as powerful as the character embedding-based model, and close to the word embedding-based model for the sentiment classification task, with much smaller vocabulary and fewer parameters. The results show that the radical embeddings are cost-effective for Chinese and Japanese. They are useful for the circumstances where the storage is limited.

There are still a lot to do on radical embeddings. For example, a radical may be related to the meaning sometimes, but express the pronunciation at other times. We will work on dealing with such phenomena for machine learning in the future.

\acks{The authors would like to thank Rakuten, Inc. and the Advanced Language Information Forum (ALAGIN) for generously providing us the Rakuten Ichiba data.}

\bibliography{acml17}






\end{document}